\documentclass[10pt]{article}
\usepackage{arxiv}
\usepackage{cite}
\usepackage[utf8]{inputenc}
\usepackage[T1]{fontenc}
\usepackage{lmodern}
\usepackage[protrusion=true,expansion=false]{microtype}
\usepackage{amsmath,amssymb,amsfonts}
\usepackage{graphicx}
\usepackage{booktabs}
\usepackage{multirow}
\usepackage{xcolor}
\usepackage{subcaption}
\usepackage{url}
\usepackage{bm}
\usepackage{nicefrac}
\usepackage{enumitem}
\usepackage{natbib}
\usepackage[colorlinks=true,linkcolor=blue,citecolor=blue,urlcolor=blue]{hyperref}
\usepackage{cleveref}

\newcommand{\orcidicon}[1]{\href{https://orcid.org/#1}{\includegraphics[width=0.32cm]{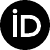}}}

\newcommand{\mol}{\textsc{MoL}}
\newcommand{\dmodel}{d_\text{model}}
\newcommand{\dthin}{d_\text{thin}}
\newcommand{\topk}{\text{top-}k}

\newcommand{\Paragraph}[1]{\vspace{0.5em}\noindent\textbf{#1}}

\title{Mixture of Layers with Hybrid Attention:\\
Parallel Thin Blocks for Sparse Transformer Compute}

\author{
  Ivan Ternovtsii \orcidicon{0009-0009-9267-8516}\thanks{This research was conducted as part of PhD studies at the Department of Software Systems, Faculty of Information Technologies, Uzhhorod National University. HengeBytes generously provided computational resources. We thank the Department of Software Systems for academic support and the HengeBytes team for maintaining the computational infrastructure. This paper reports 13 controlled experiments totaling approximately 1{,}700 GPU-hours, of which $\sim$800 GPU-hours are 1.3B-scale runs on $4{\times}$H200~NVL. Corresponding author: Ivan Ternovtsii (e-mail: ivan.ternovtsii@uzhnu.edu.ua).} \\
  Department of Software Systems, Uzhhorod National University\\
  Narodna sq. 3, Uzhhorod, Ukraine, 88000\\
  HengeBytes\\
  \texttt{ivan.ternovtsii@uzhnu.edu.ua} \\
  \And
  Yurii Bilak \orcidicon{0000-0001-5989-1643} \\
  Department of Software Systems, Uzhhorod National University\\
  Narodna sq. 3, Uzhhorod, Ukraine, 88000\\
}

\date{May 2026}

\begin{document}
\maketitle

\begin{abstract}
Standard Mixture-of-Experts (MoE) transformers route tokens to expert subnetworks within each layer, but the layer structure itself remains monolithic.
We introduce \textbf{Mixture of Layers} (\mol{}), which replaces full-width transformer blocks ($\dmodel$) with $K$ parallel \emph{thin blocks} at reduced dimensionality ($\dthin \ll \dmodel$), connected via learned down/up projections and composed via top-$k$ block routing.
Scaling sparse block routing to many blocks creates an attention coverage problem, as each block sees fewer tokens.
We address this by introducing \textbf{hybrid attention}, which pairs one shared softmax block for global context with Gated DeltaNet \citep{yang2024gated} linear attention in routed blocks.

On WikiText-103 at 85M params, MoL with dense FFN thin blocks reaches PPL~$30.95 \pm 0.11$ (3 seeds), surpassing the rank-1 expressiveness ceiling of traditional MoE by 2.98. Adding hybrid attention (1+3of15, 198M total, 77M active) reaches PPL~$29.99 \pm 0.08$ with up to $4.9\times$ forward-pass speedup from sparse dispatch. On Cosmopedia v2 (15B tokens), MoL overtakes dense baselines past 35\% of training and finishes at PPL~6.49 vs 6.65; a dense DeltaNet control at 6.64 indicates the gain is not driven by the attention swap alone.

Scaling to FineWeb-Edu 20B tokens at $d_\textrm{model}{=}2048$, \mol{} Hybrid 1+3of15 (2.08B total / 0.61B active) reaches PPL~18.04 in a single-seed run; this is 0.49~PPL better than an iso-active Dense Softmax 0.7B baseline (18.53) but 3.01~PPL worse than iso-total Dense Softmax 1.3B (15.03), and takes 1.91--2.29$\times$ longer to train than each dense baseline due to gradient-checkpointing pressure on the routed-DeltaNet kernel. Zero-shot transfer on an 8-task lm-eval-harness suite places \mol{} between the two dense baselines as capacity predicts, with a clean win on WinoGrande over Dense Softmax 1.3B (54.4 vs 51.6).
In single-GPU prefill, \mol{} Hybrid crosses Dense Softmax 1.3B at $T{\approx}5{-}6$K on the RTX~3090. On datacenter GPUs (A100, H100~SXM, H200), the crossover occurs between $T{=}64$K and $T{=}128$K. At $T{=}128$K, \mol{} ranges from near-parity ($0.98\times$ on H100~SXM) to $1.20\times$ on A100; at $T{=}256$K it reaches $1.42\times{-}1.54\times$ on H100~SXM and H200. The four measured points do not support a simple bandwidth- or compute-linear scaling, and we do not claim a hardware scaling law.
At 1.3B, \mol{} trades training efficiency and total-parameter capacity for an iso-active perplexity and downstream-transfer lead, plus a long-context prefill advantage that activates at much shorter context on the 3090 than on the three datacenter GPUs in our setup.

\end{abstract}

\section{Introduction}
\label{sec:intro}

Scaling language models traditionally means stacking more sequential transformer blocks, which increases depth, memory footprint, and latency in lockstep.
Mixture-of-Experts (MoE) transformers \citep{shazeer2017outrageously, lepikhin2021gshard, fedus2022switch, zoph2022stmoe} address parameter scaling by routing tokens to expert subnetworks, but the layer structure remains monolithic; every position runs a single full-width transformer block.

In a companion paper \citep{equifinality2026}, we found that routing topology is quality-neutral across structurally different MoE variants. If topology is not the primary driver of performance, what is? Prior work points to per-expert expressiveness: 65,536 rank-1 experts cannot replicate 256 rank-256 experts despite matched parameters (PPL~47.51 vs 12.55; see Appendix~\ref{app:rank1}).

We investigate whether layer structure itself can be decomposed, presenting four contributions:

\begin{enumerate}[nosep,leftmargin=*]
\item \textbf{Mixture of Layers (\mol{}).}
\mol{} replaces full-width transformer blocks with $K$ parallel thin blocks at $\dthin{=}256$, connected via learned down/up projections.
With dense FFNs, MoL achieves PPL~$30.95 \pm 0.11$ (3 seeds, 85M params) on WikiText-103, $2.98$ better than the full-width MoE baseline, surpassing the rank-1 expressiveness ceiling (\Cref{sec:mol}).

\item \textbf{Hybrid attention for sparse routing.}
Scaling to more blocks degrades attention quality as each block sees fewer tokens.
To address this, one shared softmax block (global coverage) is paired with Gated DeltaNet \citep{yang2024gated} routed blocks ($O(T)$ linear attention on sparse subsets).
DeltaNet outperforms softmax by 0.85~PPL after controlling for parameters in the thin-block regime ($\dthin{=}256$), an advantage that vanishes at $\dthin{=}\dmodel$ (\Cref{sec:hybrid}).

\item \textbf{Sparse dispatch with wall-clock speedup.}
Sparse dispatch (gather/scatter) yields up to $4.9\times$ forward-pass speedup at high sparsity.
DeltaNet 1+3of15 (198M total, 77M active, 3 seeds) achieves PPL~$29.99 \pm 0.08$, activating only 27\% of blocks per token (\Cref{sec:sparse}).

\item \textbf{Regime-dependent scaling.}
MoL's advantage is data-regime dependent: it loses to dense on WikiText-103 (103M tokens, multi-epoch) but wins on Cosmopedia v2 (15B tokens, single epoch), with a $0.85$~PPL swing across datasets.
A dense DeltaNet control matches dense softmax to within 0.01 PPL, indicating MoL's gain is not driven by the attention swap (\Cref{sec:scaling}).
At 1.3B-scale on FineWeb-Edu 20B tokens (single seed each), \mol{} Hybrid (2.08B total / 0.61B active, PPL~18.04) sits between iso-active Dense Softmax 0.7B (18.53) and iso-total Dense Softmax 1.3B (15.03): 0.49~PPL ahead at iso-active, 3.01~PPL behind at iso-total, with a 1.91--2.29$\times$ training-time tax that does not extend to inference (\Cref{sec:1p3b}).
\end{enumerate}

Unlike sequential hybrid designs that alternate attention types across layers \citep{qwen3next_2025, glorioso2024zamba}, \mol{} maintains full attention coverage at every layer through the shared block while confining linear attention to sparse routed subsets.

\section{The \mol{} Architecture}
\label{sec:mol}

\subsection{Thin Block Wrapper}

\mol{} replaces selected full-width transformer blocks with \textbf{split stages}, each containing $K$ independent thin blocks.
Each thin block consists of a down-projection, a standard transformer block at reduced dimensionality, and an up-projection.
\begin{equation}
\text{ThinBlock}(x) = W_\text{up} \cdot \big(\text{Block}_{\dthin}(W_\text{down} \cdot x) - W_\text{down} \cdot x\big)
\label{eq:thinblock}
\end{equation}
where $W_\text{down} \in \mathbb{R}^{\dthin \times \dmodel}$, $W_\text{up} \in \mathbb{R}^{\dmodel \times \dthin}$, and $\text{Block}_{\dthin}$ is a complete transformer block (RMSNorm $\to$ multi-head attention $\to$ RMSNorm $\to$ FFN) operating at dimensionality $\dthin$.
The subtraction strips the inner residual, so ThinBlock outputs only the delta computed by the block.

\subsection{Split Stage with Block Routing}

A split stage runs $K$ thin blocks, selects $k$ via a learned router, and averages their outputs:
\begin{equation}
\text{SplitStage}(x) = x + \frac{1}{k} \sum_{i \in \topk(K)} w_i \cdot \text{ThinBlock}_i(x)
\label{eq:splitstage}
\end{equation}
where $w_i$ are softmax-normalized routing scores and $\topk$ selects the $k$ highest-scoring blocks.
The outer residual $x + \ldots$ ensures gradient flow even if thin blocks produce small updates.
Selective block activation outperforms uniform composition by $0.99 \pm 0.10$~PPL (3 seeds, ${\sim}9\times$ pooled seed std $\sigma_\text{pooled}{=}0.12$).

\Paragraph{Load balancing.}
We use a coefficient-of-variation (CV$^2$) loss on per-block routing weights, weighted by $\alpha{=}0.05$, following standard MoE practice.
This prevents block collapse without requiring auxiliary tokens or capacity constraints.

\Paragraph{RoPE compatibility.}
All configurations maintain $d_\text{head}{=}64$: $\dmodel{=}1024$ uses 16 heads, $\dthin{=}256$ uses 4 heads.
Precomputed RoPE embeddings are shared across all block widths.

\subsection{Sparse Dispatch}
\label{sec:sparse}

With block-level routing, each thin block processes only its routed tokens.
We implement \emph{sparse dispatch}, which \textbf{gathers} only routed tokens per block, runs on compact tensors, and \textbf{scatters} results back.
This eliminates wasted compute on non-routed tokens.

\Paragraph{Correctness.}
Sparse dispatch produces numerically identical outputs to dense restricted attention (max logit diff $5.7 \times 10^{-6}$ on GPU, $2.4 \times 10^{-7}$ on CPU).

\Paragraph{Speedup.}
Forward-pass speedup scales with sparsity, reaching $1.53\times$ at 57\% active (top-4-of-7), $2.85\times$ at 20\% (top-2-of-10), and $4.34\times$ at 10\% (top-2-of-20).
With \texttt{torch.compile}: up to $4.94\times$.

\Paragraph{Projection overhead.}
Projection cost ($W_\text{down}$, $W_\text{up}$) sets a floor on $\dthin$: at $\dthin{=}256$ projections consume 40\% of wrapper parameters, rising to 57\% at $\dthin{=}128$ and 73\% at $\dthin{=}64$. Empirically $\dthin{=}256$ outperforms $\dthin{=}128$ by $1.62$~PPL at iso-parameters ($\sim$85M), so width outweighs count.

\section{Hybrid Attention}
\label{sec:hybrid}

\subsection{The Attention Coverage Problem}

Scaling to more blocks with sparse dispatch creates an attention coverage problem. At 3-of-15, each block sees only 20\% of the sequence (mean inter-token gap of 25.6), compared to 60\% at 3-of-5 (gap of 1.8).
Softmax-only sparse 3-of-15 achieves PPL~34.73 on WikiText-103, worse than 3-of-5 (32.04) despite using $2.2\times$ more parameters; more parameters cannot compensate for reduced sequence coverage.

\subsection{Architecture: Shared + Routed}

Each split stage contains:
\begin{itemize}[nosep]
\item \textbf{Shared block} (block 0): always active on all tokens, full softmax attention, providing global context at every layer.
\item \textbf{Routed blocks} (blocks 1--$N{-}1$): top-$k$ selected by the router, sparse dispatch with Gated DeltaNet \citep{yang2024gated} linear attention on their token subsets.
\end{itemize}

We denote configurations as $S{+}K\text{of}N$: $S$ shared blocks plus top-$K$ routed from $N{-}S$ candidates, with $N$ total blocks.
For example, $1{+}3\text{of}15$ has 1 shared softmax block and selects 3 from 14 routed DeltaNet blocks, activating 4 total.

\subsection{Why DeltaNet in Routed Blocks}

Gated DeltaNet is used in routed blocks; since the shared block already supplies global softmax context, the routed blocks do not require precise quadratic attention.
A $\dthin$ scaling sweep also shows DeltaNet outperforming softmax in the constrained-width regime ($\dthin{=}256$), and DeltaNet is $O(T)$ rather than $O(T^2)$, becoming faster at $T \geq 2048$.

\Paragraph{Decomposing the DeltaNet advantage.}
At $K{=}1$ (single block, no routing, all tokens visible), DeltaNet outperforms softmax by 2.55~PPL (36.28 vs 38.83) at matched architecture (${\sim}43$M params).
An iso-parameter softmax control (87.5M, $d_\text{ff}{=}1133$, 3 seeds: $32.09 \pm 0.26$) reveals that DeltaNet's extra parameters (gates, convolutions) account for 0.26~PPL (23\%) of the 1.11~PPL gap in the 1+2of5 MoL configuration; the remaining 0.85~PPL (77\%) is the parameter-controlled mechanism contribution, $3.0\times$ the pooled seed std ($\sigma_\text{pooled}{=}0.29$).

\Paragraph{Scale dependence.}
A scaling sweep across $\dthin \in \{128, 256, 512, 1024\}$ shows the DeltaNet--softmax gap shrinks monotonically; the gap is $2.79 \to 2.55 \to 1.12 \to {-}0.10$~PPL.
At $\dthin{=}\dmodel{=}1024$, softmax marginally wins ($27.60$ vs $27.70$).
DeltaNet's inductive biases (decay, delta rule) help most when attention capacity is constrained; the advantage vanishes when capacity is sufficient.

\section{Experimental Setup}
\label{sec:setup}

\subsection{Datasets}

We evaluate on three datasets of increasing scale and tokens-per-parameter:
\textbf{WikiText-103} \citep{merity2017pointer}: 103M tokens, custom BPE 32K, $T{=}255$, multi-epoch ($\sim$8 epochs); architecture ablations at 43--198M params.
\textbf{Cosmopedia v2} \citep{benallal2024cosmopedia}: 15B synthetic-textbook tokens, GPT-2 vocab, $T{=}2048$, single-epoch; data-regime dependence at $\sim$104M params.
\textbf{FineWeb-Edu} \citep{penedo2024fineweb}: filtered web tokens, GPT-2 vocab, single-epoch. 80M topology sweep at $T{=}2048$ ($\sim$2.5B tokens at 38K steps, Exp 079); 1.3B-scale runs at $T{=}4096$ (20B tokens, Exp 085/087/089).

\subsection{Architectures}

\Paragraph{80M-scale ablations} (WikiText-103, Cosmopedia): $\dmodel{=}1024$, 16 heads, 8 layers, $\dthin{=}256$, 4 heads per thin block.
Training: 50K steps (WikiText) or 114K steps (Cosmopedia), lr$=3 \times 10^{-4}$, cosine decay, 1000-step warmup (WikiText) / 2000-step (Cosmopedia), AdamW ($\beta_1{=}0.9$, $\beta_2{=}0.95$, wd$=0.01$), gradient clipping at 1.0, effective batch size 64.

\Paragraph{1.3B-scale headline} (FineWeb-Edu 20B tokens, 4$\times$ H200 NVL).
Three architectures share $d_\textrm{model}{=}2048$, 24 layers, GPT-2 50,257-vocab, $T{=}4096$, batch 1.0M tok/step, 20K steps, AdamW lr$=2\times 10^{-4}$ cosine to 0.1$\times$, 2000-step warmup, bf16 with gradient checkpointing.
\textbf{Dense Softmax 1.3B} (Exp 085): $d_\textrm{ff}{=}8192$, 16 heads ($d_\textrm{head}{=}128$). 1.31B total.
\textbf{Dense Softmax 0.7B} (Exp 089): $d_\textrm{ff}{=}2048$ matched to a single \mol{} thin block, 32 heads ($d_\textrm{head}{=}64$ matched to thin-block heads). 0.71B total.
\textbf{\mol{} Hybrid 1+3of15} (Exp 087): $\dthin{=}512$, 15 thin blocks (1 shared softmax + 14 routed DeltaNet, top-3 active = 4 active per token), $d_\textrm{ff,thin}{=}2048$. 2.08B total / 0.61B active.

\subsection{Baselines}

\begin{itemize}[nosep]
\item \textbf{Dense Softmax}: standard transformer, iso-param to MoL conditions. At 85M: PPL~30.26 (WikiText), 6.65 (Cosmopedia). At 198M: PPL~26.89 (WikiText).
\item \textbf{Dense DeltaNet}: DeltaNet attention, iso-param. At 85M: PPL~30.68 (WikiText). At 104M: PPL~6.64 (Cosmopedia), tying dense softmax within 0.01 and confirming the attention mechanism does not explain MoL's structural advantage.
\item \textbf{CT-MoE baseline} \citep{equifinality2026}: rank-1 MoE, 84.7M params, PPL~33.93 (WikiText).
\end{itemize}

\section{Results}
\label{sec:results}

\subsection{Breaking the Rank-1 Ceiling with Dense FFN Blocks}

Replacing rank-1 MoE FFNs with standard dense FFNs ($d_\text{ff} = 4 \times \dthin$) in thin blocks exceeds the rank-1 ceiling (\Cref{tab:dense_mol}).
Dense FFN MoL ($K{=}5$, top-3, 85.3M params) achieves PPL~$30.95 \pm 0.11$ (3 seeds), $2.98$ better than the full-width rank-1 baseline (33.93).

\begin{table}[t]
\centering
\caption{Dense FFN thin blocks vs rank-1 MoE on WikiText-103. All at $\dthin{=}256$, 50K steps.}
\label{tab:dense_mol}
\small
\begin{tabular}{lcccc}
\toprule
Config & FFN & Params & PPL & $\Delta$ vs CT-MoE \\
\midrule
CT-MoE baseline (rank-1) & MoE & 84.7M & 33.93 & --- \\
MoL K=7 split\_all (rank-1) & MoE & 85.6M & 34.83 & $+0.90$ \\
\midrule
\textbf{MoL K=5 top-3 (dense)} & Dense & 85.3M & $\mathbf{30.95 \pm 0.11}$ & $\mathbf{-2.98}$ \\
MoL K=5 all-active (dense) & Dense & 85.3M & $31.94 \pm 0.12$ & $-1.99$ \\
\bottomrule
\end{tabular}
\end{table}

\subsection{Hybrid Attention Results}

\Cref{tab:hybrid_results} summarizes the hybrid attention experiments on WikiText-103.
DeltaNet 1+2of5 ($31.24 \pm 0.31$, 3 seeds) outperforms both iso-param Softmax 1+2of5 ($32.09 \pm 0.26$) and standard Softmax 1+2of5 ($32.35 \pm 0.15$).
Scaling to 15 blocks, DeltaNet 1+3of15 (198.2M total, 77M active) achieves PPL~$\mathbf{29.99 \pm 0.08}$ (3 seeds), activating only 27\% of blocks per token.

\begin{table}[t]
\centering
\caption{Hybrid attention on WikiText-103. S=shared (all tokens), R=routed (sparse). $\dagger$=1 seed; $\pm$=3 seeds.}
\label{tab:hybrid_results}
\small
\begin{tabular}{llcccc}
\toprule
Config & Attn & Active & Params & PPL \\
\midrule
Dense full-attn K=5$^\dagger$ & softmax & 3/5 & 85.3M & 30.85 \\
Sparse 3-of-5 & softmax & 3/5 & 85.3M & $32.19 \pm 0.15$ \\
\midrule
Softmax 1+2of5 iso-param & softmax & 1S+2R & 87.5M & $32.09 \pm 0.26$ \\
\textbf{DeltaNet 1+2of5} & S:smx, R:$\Delta$ & 1S+2R & 87.5M & $\mathbf{31.24 \pm 0.31}$ \\
\midrule
Softmax 1+2of15$^\dagger$ & softmax & 1S+2R & 190.4M & 36.51 \\
DeltaNet 1+2of15$^\dagger$ & S:smx, R:$\Delta$ & 1S+2R & 198.2M & 32.75 \\
\textbf{DeltaNet 1+3of15} & S:smx, R:$\Delta$ & 1S+3R & 198.2M & $\mathbf{29.99 \pm 0.08}$ \\
\midrule
Dense softmax 85M$^\dagger$ & softmax & --- & 84.7M & 30.26 \\
Dense softmax 198M$^\dagger$ & softmax & --- & 198.2M & 26.89 \\
Dense DeltaNet 85M$^\dagger$ & DeltaNet & --- & 84.7M & 30.68 \\
\bottomrule
\end{tabular}
\end{table}

\Paragraph{Dense baselines contextualize MoL quality.}
At iso-total-parameters (198M), a dense softmax transformer reaches PPL~26.89, $3.1$~PPL better than DeltaNet 1+3of15. This establishes the cost of sparse block routing.
Note that \mol{} activates only 77M parameters per token (39\% of total), so this gap is paid against a model with $2.6\times$ fewer per-token FLOPs and admits block-parallel placement across devices.

\subsection{Data-Regime Scaling: Cosmopedia 15B}
\label{sec:scaling}

On WikiText-103 (103M tokens, ${\sim}$8 epochs), MoL consistently underperforms dense: MoL K=5 top-3 (85.3M, $30.95 \pm 0.11$) loses by 0.69~PPL vs dense softmax ($30.26$).
Training on Cosmopedia v2 (15B unique tokens, single epoch) reverses this (\Cref{tab:cosmopedia}).

\begin{table}[t]
\centering
\caption{Cosmopedia v2 results (15B tokens, single epoch, ${\sim}$104M params). MoL overtakes dense at step 40K.}
\label{tab:cosmopedia}
\small
\begin{tabular}{lccc}
\toprule
Condition & Params & Final PPL & vs Dense \\
\midrule
Dense softmax & 103.4M & 6.65 & --- \\
Dense DeltaNet & 103.4M & 6.64 & $-0.01$ \\
\textbf{MoL K=5 top-3} & 104.0M & \textbf{6.49} & $\mathbf{-0.16}$ \\
Sparse MoL 1+2of5 & 106.0M & 6.52 & $-0.13$ \\
MoE-FFN 3of5 & ${\sim}$106M & 6.89 & $+0.24$ \\
\bottomrule
\end{tabular}
\end{table}

MoL K=5 crosses the dense baseline at step 40K (35\% through training) and stabilizes at $-0.16$~PPL, representing a \textbf{0.85~PPL swing} relative to WikiText-103.
The sparse MoL variant (DeltaNet routed + softmax shared) tracks within 0.03~PPL (6.52 vs 6.49), retaining 81\% of the MoL gain under sparse dispatch.

\Paragraph{Isolating the structural contribution.}
Dense DeltaNet (103.4M, Cosmopedia) finishes at PPL~6.64, within 0.01 of dense softmax (6.65).
Since both dense baselines converge to the same PPL, MoL's 0.15--0.16~PPL lead cannot be attributed to the linear attention choice.
The advantage derives from block-level routing with joint attention+FFN specialization.
An MoE-FFN ablation (shared attention + routed FFN experts, no block routing) finishes at PPL~6.89 ($+0.24$ vs dense), confirming that FFN-only routing does not explain MoL's gain.

\subsection{FineWeb-Edu 80M Ablations}
\label{sec:fineweb_80m}

We test whether the \mol{} advantage on Cosmopedia (\Cref{sec:scaling}) replicates on web-text by running an 80M-scale sweep on FineWeb-Edu for 60K steps ($T{=}2048$, $\sim$3.9B tokens, 3$\times$ local GPUs, single seed each), comparing three \mol{} Hybrid topologies (1+3of\{5, 10, 15\}) against two Dense Softmax baselines bracketing both ends of the active-parameter spectrum (\Cref{tab:fineweb_80m}).

\begin{table}[h]
\centering
\caption{FineWeb-Edu 80M sweep at 60K steps (Exp 079). MoL conditions use shared softmax and routed Gated DeltaNet ($\dthin{=}256$). Differences are relative to the Dense 98M baseline. Single seed each.}
\label{tab:fineweb_80m}
\small
\begin{tabular}{lrrrr}
\toprule
Condition & Total & Active & Val PPL @ 60K & $\Delta$ PPL \\
\midrule
Dense Softmax 152M (iso-FLOP)        & 152M  & 152M  & \textbf{24.97} & $-4.61$ \\
\textbf{Dense Softmax 98M (iso-active)} & 97.7M & 97.7M & \textbf{29.58} & --      \\
\mol{} Hybrid 1+3of15                 & 217M  & 95M   & \textbf{26.40} & $-3.18$ \\
\mol{} Hybrid 1+3of10                 & 162M  & 95M   & 27.76          & $-1.82$ \\
\mol{} Hybrid 1+3of5                  & 106M  & 95M   & 27.96          & $-1.62$ \\
\bottomrule
\end{tabular}
\end{table}

\textbf{At iso-active compute} (Dense 98M with $d_\textrm{ff}{=}768$ matched to MoL's $\sim$95M active), all three \mol{} topologies cross Dense between steps 10K--12K and finish 1.62--3.18~PPL ahead by 60K, with the gap widening through training.
\textbf{At iso-FLOP} (Dense 152M, $1.6\times$ more active compute), Dense finishes 1.43~PPL ahead of best \mol{}, confirming that \mol{} trades total parameters for active-compute efficiency, the same trade as MoE but at the block level. A $K_\textrm{active}$ sweep at fixed pool $N{=}5$ (\Cref{app:kactive}) shows the result holds across different block counts.

\subsection{1.3B-Scale Results}
\label{sec:1p3b}

We trained three architectures on FineWeb-Edu 20B tokens to compare \mol{} Hybrid against Dense Softmax baselines bracketing it on both ends of the active-parameter spectrum.

\begin{table}[h]
\centering
\caption{1.3B-scale validation perplexity on FineWeb-Edu 20B tokens (4$\times$ H200 NVL, identical optimizer / token budget). Wall time is the full 20K-step run.}
\label{tab:1p3b_headline}
\small
\begin{tabular}{lrrrrr}
\toprule
Model & Total & Active & $d_\textrm{ff}$ & Final PPL & Wall time \\
\midrule
Dense Softmax 1.3B (Exp 085)        & 1.31B & 1.31B & 8192 & \textbf{15.03} & 53.3h \\
Dense Softmax 0.7B (Exp 089)        & 0.71B & 0.71B & 2048 & 18.53 & 44.5h \\
\mol{} Hybrid 1+3of15 (Exp 087)     & 2.08B & 0.61B & 2048$^\dagger$ & 18.04 & 102.0h \\
\bottomrule
\end{tabular}
\\[2pt]
\raggedright\footnotesize $^\dagger$ Per thin block; 4 active blocks per token.
\end{table}

\paragraph{Iso-active comparison.} On the closest active-parameter match, \mol{} Hybrid (0.61B active) reaches 18.04 PPL vs Dense Softmax 0.7B (0.71B active, $d_\textrm{ff}$ matched to a single thin block) at 18.53, a $0.49$~PPL gap; \mol{} crosses Dense 0.7B at step~7K and stabilizes at $\sim$1.03$\times$ ratio. The 1.3B comparison uses one seed per condition; cross-seed std on the same architecture is $\sigma{\approx}0.08$~PPL on WikiText-103 (\Cref{app:multiseed}).

\begin{figure}[t]
\centering
\includegraphics[width=0.75\columnwidth]{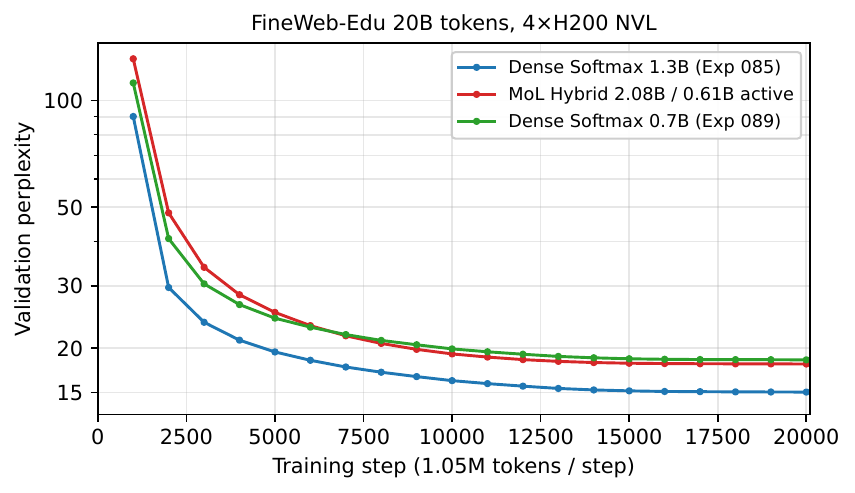}
\caption{Validation perplexity (log scale) vs training step for the three 1.3B-scale runs on FineWeb-Edu 20B tokens.
Dense Softmax 1.3B (blue) leads from step~1; \mol{} Hybrid (red) stays $\sim$3~PPL above through training; Dense Softmax 0.7B (green) starts above \mol{} and crosses below at step~7K, then narrows the gap from $-3.5$~PPL to $-0.49$~PPL by step~20K.
The Dense 1.3B vs \mol{} gap stabilizes at step~11K and does not close within budget.}
\label{fig:loss_1p3b_traj}
\end{figure}

\paragraph{Iso-total gap.} Dense Softmax 1.3B finishes 3.01~PPL ahead of \mol{}'s 2.08B total; the gap stabilizes at step~11K (\Cref{fig:loss_1p3b_traj}). Candidate causes (active FFN capacity, within-\mol{} sparsity, sub-Chinchilla token budget) are discussed in \Cref{app:isototal}.

\paragraph{Zero-shot transfer.} On an 8-task lm-eval-harness suite (7 tasks in \Cref{tab:1p3b_zeroshot}; TruthfulQA mc1 in \Cref{app:zeroshot}), \mol{} wins 4 of 8 tasks against iso-active Dense~0.7B (HellaSwag +1.0, ARC-Challenge +0.6, PIQA +0.1, WinoGrande +5.0) and edges Dense~1.3B on WinoGrande (54.4 vs 51.6). \mol{} trails Dense~1.3B on the broad-knowledge tasks (HellaSwag, ARC-Easy, PIQA) by 1.5--4.9~pts, consistent with the iso-total PPL gap, and trails iso-active Dense~0.7B on TruthfulQA mc1/mc2 (a documented inverse-scaling regime). All three sit at chance on MMLU 5-shot ($\sim$24\%), expected at 1.3B~/~20B tokens; the pattern places \mol{}'s downstream transfer at fixed active compute between the density-matched dense baselines, as capacity predicts.

\begin{table}[h]
\centering\small
\caption{Zero-shot accuracy (\%, lm-eval-harness 0.4.7, bf16; \texttt{acc\_norm} where benchmark prefers it; MMLU 5-shot, others 0-shot). Best per row in bold; MMLU rows are within noise of chance ($\sim$25\%) so we omit bolding there. Stderr per task is $0.005$--$0.014$.}
\label{tab:1p3b_zeroshot}
\begin{tabular}{lccc}
\toprule
Task & Dense 1.3B (085) & Dense 0.7B (089) & \mol{} 0.61B/2.08B (087) \\
\midrule
MMLU (5-shot)    & 24.31 & 23.98 & 23.07 \\
HellaSwag        & \textbf{39.71} & 34.10 & 35.10 \\
ARC-Easy         & \textbf{53.07} & 48.86 & 48.15 \\
ARC-Challenge    & \textbf{26.88} & 25.43 & 26.02 \\
PIQA             & \textbf{65.56} & 63.98 & 64.09 \\
WinoGrande       & 51.62 & 49.33 & \textbf{54.38} \\
TruthfulQA mc2   & 37.83 & \textbf{40.89} & 39.55 \\
\bottomrule
\end{tabular}
\end{table}

\paragraph{Training cost.} \mol{} Hybrid takes $1.91\times$ Dense 1.3B's wall time (102.0h vs 53.3h) at identical token budget. Two costs compound. First, gradient checkpointing is mandatory at 2.08B total params on 141\,GB H200s. Second, DeltaNet's chunked kernel is more expensive to replay during backward than FlashAttention. On Cosmopedia 80M (no checkpointing), the same architecture was $1.45\times$ \emph{faster} than Dense (\Cref{app:speed_crossover}); the training tax is scale- and memory-budget-specific, not intrinsic.

\paragraph{Inference inversion at long context.} The training penalty inverts at inference (single-GPU, batch=1, prefill). On RTX~3090, \mol{} Hybrid is $1.25\times{-}1.76\times$ faster than Dense Softmax 1.3B at $T{=}8{-}32$K. On the three datacenter GPUs we measured (A100, H100~SXM, H200), the crossover sits between $T{=}64$K and $T{=}128$K, with \mol{} ranging from near-parity ($0.98\times$ on H100~SXM) to $1.20\times$ on A100 at $T{=}128$K and $1.42\times{-}1.54\times$ at $T{=}256$K (\Cref{sec:inference}). At $T{=}256$K, \mol{}~Hybrid 2.08B successfully completes prefill on A100~80GB whereas Dense Softmax 1.3B OOMs, a memory-pressure manifestation of \mol{}'s smaller per-layer softmax KV cache. The four measured points do not support a clean bandwidth- or compute-linear scaling story; we report the empirical crossovers and leave a multi-architecture mechanistic model to future work. At 1.3B, \mol{} trades a training premium and an iso-total PPL gap for an iso-active PPL and downstream-transfer lead, plus a long-context prefill advantage whose onset is hardware-specific.

\subsection{Inference Wall-Clock: Prefill and Decode}
\label{sec:inference}

We benchmark prefill and decode throughput at batch=1 on two GPU classes (bf16, post-warmup, FlashAttention-2 for softmax, FLA Triton kernels for DeltaNet, KV cache for both Dense and \mol{}): a memory-bandwidth-limited \textbf{RTX 3090} (936 GB/s HBM, 71~TFLOPS bf16) and a compute/bandwidth-rich \textbf{H200} (4.8 TB/s HBM, 989~TFLOPS bf16). For the 3090 prefill table we add two control architectures to isolate the routing contribution from the attention-type contribution: a \emph{Dense DeltaNet 1.3B} (iso-architecture to Dense Softmax with the attention swapped) and a \emph{\mol{} all-softmax 2.08B} (same routing as \mol{} Hybrid but with \texttt{routed\_attn\_type=softmax}).

\subsubsection{Prefill: Long Context}

\begin{table}[h]
\centering
\caption{Prefill throughput (tokens/s, batch=1, single GPU; 5-run mean per cell after warmup, run-to-run variance ${<}2\%$ at fixed $T$ on each GPU). \mol{} Hybrid value is the faster of \texttt{batched\_sparse} and \texttt{sparse} at each $T$ (both shown where measured; see protocol note). \emph{Best per row in bold} marks the crossover between Dense and \mol{} Hybrid. The four measured GPUs do not show a clean monotonic shift in crossover position with HBM bandwidth or compute; we present empirical positions only.}
\label{tab:inference_long_ctx}
\small
\begin{tabular}{rrrrr}
\toprule
SeqLen & Dense Softmax 1.3B & Dense DeltaNet 1.3B$^\dagger$ & \mol{} all-softmax 2.08B$^\dagger$ & \mol{} Hybrid 2.08B \\
\midrule
\multicolumn{5}{l}{\emph{RTX 3090 (consumer; 936 GB/s HBM, 71 TFLOPS bf16)}}\\
\;\;4{,}096   & \textbf{18{,}041} & 17{,}962 & 20{,}591 & 16{,}005 \\
\;\;32{,}768  & 10{,}808 & 20{,}502 & 16{,}424 & \textbf{19{,}062} \\
\addlinespace
\multicolumn{5}{l}{\emph{A100 SXM4 80GB (1.55 TB/s HBM, 312 TFLOPS bf16)}}\\
\;\;4{,}096    & \textbf{43{,}585} & — & — & 9{,}998 \\
\;\;32{,}768   & \textbf{28{,}472} & — & — & 16{,}650 \\
\;\;65{,}536   & \textbf{19{,}495} & — & — & 15{,}968 \\
\;\;131{,}072  & 11{,}779 & — & — & \textbf{14{,}115} (sparse) \\
\addlinespace
\multicolumn{5}{l}{\emph{H100 SXM 80GB (3.35 TB/s HBM, 989 TFLOPS bf16)}}\\
\;\;4{,}096    & \textbf{98{,}476} & — & — & 16{,}459 \\
\;\;32{,}768   & \textbf{71{,}925} & — & — & 33{,}579 \\
\;\;65{,}536   & \textbf{51{,}731} & — & — & 33{,}559 \\
\;\;131{,}072  & \textbf{31{,}299} & — & — & 30{,}819 (sparse, ${\approx}0.98\times$ Dense) \\
\;\;262{,}144  & 17{,}495 & — & — & \textbf{24{,}841} (sparse) \\
\addlinespace
\multicolumn{5}{l}{\emph{H200 (4.8 TB/s HBM, 989 TFLOPS bf16)}}\\
\;\;4{,}096    & \textbf{102{,}460} & — & — & 14{,}174 \\
\;\;32{,}768   & \textbf{71{,}912} & — & — & 36{,}911 \\
\;\;65{,}536   & \textbf{51{,}860} & — & — & 36{,}292 (sparse) / 35{,}244 (batched) \\
\;\;131{,}072  & 31{,}348 & — & — & \textbf{33{,}008} (sparse) \\
\;\;262{,}144  & 16{,}973 & — & — & \textbf{26{,}105} (sparse) \\
\bottomrule
\end{tabular}
\\[2pt]
\raggedright\footnotesize $^\dagger$ Random-initialized control architectures (not trained); used on 3090 to attribute the speedup to routing vs attention-type. \textbf{Path-selection protocol}: \texttt{batched\_sparse} stacks all $N{=}14$ routed FLA calls into one kernel (peak $\sim$85\,GB at 128K) and dominates short context where dispatch overhead matters; \texttt{sparse} dispatches per block (peak $\sim$32\,GB at 128K) and dominates long context where the stacked footprint OOMs. We report the faster path at each $T$. \textbf{Bench limits}: Dense Softmax 1.3B OOMs above 524K (H200) and 128K (A100/H100 SXM 80GB); \mol{} sparse OOMs above 262K with the current KV-cache implementation; FlashAttention-3, fp8 KV-cache, or multi-GPU serving could move all ceilings.
\end{table}

\paragraph{Empirical crossover positions across four GPUs.} On RTX 3090, \mol{} Hybrid crosses Dense Softmax at $T{\approx}5{-}6$K and reaches $1.76\times$ Dense at $T{=}32$K. On the three datacenter GPUs we measured (A100 80GB, H100 SXM 80GB, H200 144GB), the crossover sits between $T{=}64$K and $T{=}128$K despite a 3$\times$ HBM-bandwidth spread (1.55$\to$4.8 TB/s) and a 3$\times$ compute spread (312$\to$989 TFLOPS bf16): \mol{} reaches $1.20\times$ Dense at $T{=}128$K on A100, ${\approx}1.0\times$ on H100 SXM, $1.05\times$ on H200, and $1.42\times{-}1.54\times$ Dense at $T{=}256$K on H100~SXM and H200 (Dense OOMs at $T{=}256$K on A100~80GB while \mol{}~Hybrid 2.08B does not, isolating the memory advantage of \mol{}'s smaller per-layer softmax KV cache). \emph{The four measured points do not support a simple monotonic scaling with HBM bandwidth or compute alone.} Plausible contributing factors include differences in FlashAttention kernel efficiency across architectures, GDDR6X vs HBM memory hierarchies, L2 cache sizes, and SM-to-bandwidth ratios; we do not isolate these and we do not claim a closed-form scaling law. We report the empirical crossovers above as observed in our single-GPU, batch=1, FA-2/FLA bench. Our earlier H200 sweep stopped at 32K and led us to incorrectly conclude no crossover existed on Hopper; the 4-hardware data shows the crossover is universal in our measurements but late on datacenter GPUs.

\paragraph{Decomposing the 3090 speedup.}
The per-token latency reduction on 3090 at $T{=}32$K (Dense Softmax $92.5\,\mu$s $\to$ \mol{} Hybrid $52.5\,\mu$s) decomposes into a routing-structure contribution of $31.6\,\mu$s ($\mathbf{79\%}$, Dense Softmax $\to$ \mol{} all-softmax: smaller per-block $d_\textrm{ff}$ and 14 narrow routed-attention windows of $\sim T/4.67$ tokens vs one giant $T{\times}T$) and a DeltaNet-on-routed contribution of $8.4\,\mu$s ($\mathbf{21\%}$, \mol{} all-softmax $\to$ \mol{} Hybrid: $O(T)$ linear attention on the routed pathway). Dense DeltaNet 1.3B reaches $1.90\times$ Dense Softmax at $T{=}32$K on 3090 (the $O(T)$ vs $O(T^2)$ effect alone). Both decomposition runs are 3090-only; an H200 decomposition would require re-running the controls and is left for future work.

\paragraph{Implementation.} The numbers above use the production inference path \texttt{attention\_mode="batched\_sparse"}, which fuses the $N{=}14$ routed FLA calls into one \texttt{chunk\_gated\_delta\_rule} call with $N{\times}H$ heads and stacks per-block linear projections via \texttt{bmm}. The naive (sparse) path is $2.8\times$ slower than batched on 3090 at $T{=}256$ batch=1, and $2.77\times$ slower on H200 at $T{=}4$K (consistent across hardware). Numerical equivalence is preserved: validation perplexity on the Exp 087 trained checkpoint shifts only $-0.11\%$ ($18.48 \to 18.46$, within bf16 noise).

\paragraph{Decode (3090 KV-cache).} On 3090 with KV cache enabled, Dense Softmax decode latency grows linearly at $\sim$3~ms per 4K context (6.2~ms/tok @256 $\to$ 26.5~ms/tok @24K, $4.3\times$), while \mol{} stays flat at $\sim$60--65~ms across the same range, dominated by a per-token dispatch floor of $\sim$2,160 Python op launches; the asymptotic flat-vs-linear property holds but the floor makes \mol{} slower than Dense in absolute terms across all measured contexts. Detailed decode tables and the H200 decode follow-up are deferred to \Cref{app:decode}.

\paragraph{Caveats.} The Dense DeltaNet 1.3B prefill control uses random weights (Exp 086 was planned but not trained, so we attribute wall-clock without claiming joint quality+speed); at $T\leq 2$K batch=1 on 3090, Dense Softmax still wins by up to $0.78\times$ on prefill.

\subsection{Limitations}
\label{sec:limitations}

Four limitations scope our claims.
\textbf{(1)} 1.3B uses single-seed runs; cross-seed $\sigma{\approx}0.08$~PPL on 80M WikiText-103 (\Cref{app:multiseed}) does not bound scale variance.
\textbf{(2)} \mol{}'s 1.91--2.29$\times$ training cost at 1.3B reflects gradient-checkpointing pressure on the routed-DeltaNet kernel; at 80M Cosmopedia (no checkpointing) the same architecture was $1.45\times$ \emph{faster} than Dense, so the tax is memory-budget-specific.
\textbf{(3)} Prefill advantage is single-GPU, batch=1 on FlashAttention-2 + FLA kernels, with hardware-specific onset ($T{\approx}5{-}6$K on RTX~3090; $T{=}64$--$128$K on A100/H100/H200; \Cref{sec:inference}). Multi-GPU parallelism, batched serving, FlashAttention-3, and fp8 KV-cache could shift the crossover; we report a controlled microbenchmark, not a deployment result. \mol{} decode is slower at every measured context (3.1--9.7$\times$) and we do not extrapolate beyond 24K.
\textbf{(4)} The 1.3B-scale comparison uses FineWeb-Edu 20B tokens at $\sim$15.6~tokens/active-param; larger budgets and additional corpora are needed to test whether the iso-total gap persists.

\section{Related Work}
\label{sec:related}

\Paragraph{MoE and layer modularity.}
Standard MoE \citep{shazeer2017outrageously, fedus2022switch, zoph2022stmoe} routes tokens to FFN experts within a full-width layer. In contrast, \mol{} routes over whole transformer blocks at reduced $\dmodel$. Layer-level modularity work modifies \emph{which} layers execute (MoD \citep{raposo2024mod}, MoLEx \citep{teo2025molex}, MoR \citep{bae2025mor}, MoEUT \citep{csordas2024moeut}); sub-network parallelism partitions weights (UoE \citep{yang2025uoe}), the FFN only (FlashMHF \citep{zhang2026flashmhf}), or sub-blocks (AltUp \citep{baykal2023altup}); MoM \citep{gong2024mom} and MoUE \citep{chen2026moue} pool/share modules, all at full $\dmodel$. No prior work combines parallel blocks, reduced-$\dmodel$ down/up projections (used as fine-tuning in adapters/LoRA \citep{houlsby2019adapters,hu2022lora}), top-$k$ block routing, and hybrid attention.

\Paragraph{Hybrid attention.}
Unlike sequential hybrids that alternate attention types across layers \citep{qwen3next_2025, glorioso2024zamba, gu2024mamba, peng2023rwkv}, \mol{} carries softmax (shared) and $O(T)$ DeltaNet \citep{yang2024gated} (routed) \emph{within} every layer.

\bibliographystyle{plainnat}
\bibliography{references}

\appendix

\section{The Rank-1 Expressiveness Ceiling}
\label{app:rank1}

Before decomposing layer structure, we established the fundamental quality constraint: per-expert expressiveness.
A rank-1 mirror experiment compares 65,536 rank-1 experts against 256 rank-256 experts at matched total expert parameters (${\sim}67$M/layer) and matched active compute ($K{=}1024$ rank-1 vs $K{=}4$ rank-256).

\begin{table}[h]
\centering
\caption{Rank-1 mirror experiment. 65,536 rank-1 experts cannot replicate 256 rank-256 experts.}
\label{tab:rank1_mirror}
\small
\begin{tabular}{lcccc}
\toprule
Model & $N$/layer & Rank & Params & PPL \\
\midrule
High-rank reference & 256 & 256 & ${\sim}$290M & \textbf{12.55} \\
Rank-1 mirror & 65,536 & 1 & 322.7M & 47.51 \\
CT-MoE Wide 1$\times$12 & 1,024 & 1 & 84.7M & 33.93 \\
\bottomrule
\end{tabular}
\end{table}

The rank-1 mirror achieves PPL~47.51, $3.8\times$ worse, despite 11\% more total parameters.
This establishes that the rank-1 constraint is the binding quality limit: architectural rearrangements operate within this ceiling, motivating the transition to dense FFN thin blocks (\Cref{sec:mol}).

\section{Active Compute Scaling}
\label{app:active_compute}

Scaling from $K{=}12$ to $K{=}128$ active rank-1 experts ($10\times$ more active FLOPs, $N{=}4096$, 138M params, $\tau{=}15$) yields PPL~33.22, closing only 19\% of the 3.67-PPL gap to the dense baseline (30.26) despite $10\times$ more active computation.
This confirms that quality is limited by \emph{per-expert expressiveness}, not the number of active experts.

\section{Activation-Level Gating}
\label{app:se_gating}

Squeeze-and-Excitation (SE) gating \citep{hu2018squeeze} applied to expert activations (not $\dmodel$-dimensional vectors) achieves PPL~31.82, the best iso-parameter rank-1 result.
A small MLP ($K {\to} 4 {\to} K$, 896 extra parameters total) modulates routing weights based on expert activation magnitudes.
SE closes 46\% of the dense gap (33.93 $\to$ 31.82, target 29.36).
Neither active compute scaling nor activation gating fully close the gap, motivating the structural decomposition of \mol{}.

\section{MoL Granularity Sweep}
\label{app:granularity}

\begin{table}[h]
\centering
\caption{MoL granularity sweep on WikiText-103 (50K steps). All with rank-1 MoE FFN.}
\label{tab:mol_sweep}
\small
\begin{tabular}{lccccc}
\toprule
Config & $\dthin$ & $K$ & Params & PPL & Speed \\
\midrule
CT-MoE baseline & 1024 & --- & 84.7M & 33.93 & --- \\
K=7 d=256 split\_all & 256 & 7 & 85.6M & \textbf{34.83} & 1.51 s/s \\
K=7 d=256 split\_mid6 & 256 & 7 & 85.4M & 35.05 & 1.85 s/s \\
K=18 d=128 & 128 & 18 & 86.1M & 36.67 & 1.08 s/s \\
K=40 d=64 & 64 & 40 & 84.7M & 38.65 & 0.69 s/s \\
K=1 d=256 & 256 & 1 & 51.4M & 39.71 & 4.87 s/s \\
\bottomrule
\end{tabular}
\end{table}

$K{=}7$ at $\dthin{=}256$ with all layers split reaches PPL~34.83, only 0.90 worse than the baseline.
Width matters more than count at iso-parameters: $\dthin{=}256$ ($K{=}7$) outperforms $\dthin{=}128$ ($K{=}18$) by 1.62~PPL.
Full-width anchor layers are unnecessary: split\_all (34.83) outperforms split\_mid6 (35.05).

\section{DeltaNet Speed Crossover}
\label{app:speed_crossover}

\begin{table}[h]
\centering
\caption{DeltaNet vs softmax attention speed (isolated block, forward+backward).}
\label{tab:speed_crossover}
\small
\begin{tabular}{rrrr}
\toprule
SeqLen & Softmax (ms) & DeltaNet (ms) & Ratio \\
\midrule
512 & 1.07 & 3.12 & 0.34$\times$ \\
1024 & 1.89 & 3.18 & 0.59$\times$ \\
\textbf{2048} & \textbf{4.04} & \textbf{3.33} & $\mathbf{1.21\times}$ \\
4096 & 11.06 & 4.81 & 2.30$\times$ \\
8192 & 36.08 & 8.93 & 4.04$\times$ \\
16384 & 131.81 & 16.88 & 7.81$\times$ \\
\bottomrule
\end{tabular}
\end{table}

DeltaNet crosses softmax at $T{\approx}2048$ and reaches $7.8\times$ faster at $T{=}16$K.

\section{Block-Parallel Model Sharding (Analytic)}
\label{app:parallelism}

\mol{}'s thin blocks have no within-layer cross-block data dependencies, which makes them a natural fit for expert-parallel placement.
A prototype distributing 14 routed blocks across 3 GPUs over PCIe (with differentiable all-to-all token dispatch) yields a 38\% per-GPU parameter reduction and 23\% VRAM savings at a 34\% wall-clock penalty; the penalty is expected to shrink substantially on NVLink.
Compared to Megatron-style tensor parallelism, which incurs two all-reduces per transformer layer in the forward pass \citep{narayanan2021efficient}, and to standard expert-parallel MoE, where dispatch/combine collectives per MoE layer can dominate step time at scale \citep{deepseekv3}, \mol{}'s block-parallel design requires only one all-reduce per split stage and keeps attention local to each block.
A larger-scale strategy would combine block placement (primary axis), pipeline depth, and context parallelism inside long routed sub-sequences; we leave a measured throughput study to future work.

\section{Multi-Seed Validation}
\label{app:multiseed}

\begin{table}[h]
\centering
\caption{Multi-seed validation (seeds 42, 137, 256) on WikiText-103, 50K steps.}
\label{tab:multiseed}
\small
\resizebox{\columnwidth}{!}{%
\begin{tabular}{lcccccc}
\toprule
Seed & 1+3of15 & 1+2of5 & Softmax 1+2of5 & Sparse 3/5 & Dense top-3 & Dense all \\
 & (198M) & (87.5M) & (85.3M) & (85.3M) & (85.3M) & (85.3M) \\
\midrule
42 & 29.96 & 30.95 & 32.46 & 32.04 & 30.85 & 31.80 \\
137 & 29.94 & 31.57 & 32.41 & 32.33 & 30.93 & 32.03 \\
256 & 30.08 & 31.20 & 32.19 & 32.21 & 31.06 & 31.98 \\
\midrule
Mean$\pm$std & $29.99 \pm 0.08$ & $31.24 \pm 0.31$ & $32.35 \pm 0.15$ & $32.19 \pm 0.15$ & $30.95 \pm 0.11$ & $31.94 \pm 0.12$ \\
\bottomrule
\end{tabular}%
}
\end{table}

\section{Decode Latency: 3090 KV-cache Tables}
\label{app:decode}

The decode bench in \Cref{sec:inference} reports a single 3090 summary line; the full table is reproduced here. \mol{} exhibits \emph{sub-linear} decode latency growth with context: Dense Softmax has $O(T)$ KV-cache memory traffic on every layer's attention, while \mol{} splits attention across a shared softmax block (also $O(T)$) and routed DeltaNet blocks ($O(1)$ recurrent state). With 1 shared and 3 routed-active per token in 1+3of15, only $\sim$1/4 of attention work is $O(T)$, so decode latency grows much more slowly with context than Dense.

\begin{table}[h]
\centering\small
\caption{Decode latency per token (ms, batch=1, RTX 3090, KV cache enabled). Dense scales linearly; \mol{} stays flat at $\sim$60ms, dominated by a per-block Python dispatch floor that masks the smaller linear-growth term from \mol{}'s single shared softmax block.}
\label{tab:decode_latency}
\begin{tabular}{rrrr}
\toprule
Context & Dense Softmax 1.3B (ms/tok) & \mol{} Hybrid 2.08B (ms/tok) & Dense growth \\
\midrule
256     & \textbf{6.2}  & 63.2 & — \\
1{,}024 & \textbf{6.9}  & 64.2 & +12\% \\
4{,}096 & \textbf{9.6}  & 63.3 & +55\% \\
8{,}192 & \textbf{12.9} & 62.5 & +108\% \\
16{,}384 & \textbf{19.6} & 65.2 & +216\% \\
24{,}576 & \textbf{26.5} & $\sim$65 & +328\% \\
\bottomrule
\end{tabular}
\end{table}

The asymptotic property is measured directly: \emph{Dense's decode latency grows $4.3\times$ over 256--24K while \mol{}'s grows $<5\%$.} RTX~3090 memory caps direct measurement at $\sim$24K, so we do not extrapolate beyond. KV cache alone gives $2.74\times$ / $1.92\times$ / $6.19\times$ uplift on Dense 1.3B / Dense 0.7B / \mol{} at short context. The $\sim$60~ms floor would shrink under a per-block-fused kernel: a partial fix (FLA-only fusion) gave $+8\%$ throughput. Full fusion is deferred. The decode bench was not re-run on H200; the prefill measurements (\Cref{sec:inference}) show the same crossover regime appears on Hopper at $T{\approx}96{-}128$K (a $\sim$25$\times$ shift vs 3090), so the asymptotic decode-latency story should hold on H200 starting from a higher absolute threshold.

\section{$K_\textrm{active}$ Sweep at 80M (Fixed Pool $N{=}5$)}
\label{app:kactive}

To isolate the effect of $K_\textrm{active}$ at fixed pool size, we ran three additional 60K conditions on FineWeb-Edu, varying $K_\textrm{active}$ while applying the \mol{} design rule $K_\textrm{active} \times d_\textrm{expert} \approx d_\textrm{model}$ and $K_\textrm{active} \times d_\textrm{ff,thin} \approx d_\textrm{ff,dense}$ (active FFN width held at $\sim$4096):

\begin{table}[h]
\centering\small
\caption{$K_\textrm{active}$ sweep at 80M, $N{=}5$, FineWeb-Edu 60K steps. All conditions iso-active-FFN-width $\sim$4096. Single seed each.}
\label{tab:kactive}
\begin{tabular}{lrrrrrrr}
\toprule
Condition & $K$ & $d_\textrm{expert}$ & $d_\textrm{ff,thin}$ & Total & Active & PPL @ 60K & $\Delta$ vs Dense 98M \\
\midrule
1+1of5 (K=2)            & 2 & 512 & 2048 & 228M & 119M & \textbf{27.47} & $-2.11$ \\
1+2of5 (K=3)            & 3 & 320 & 1344 & 132M &  98M & 27.51 & $-2.07$ \\
1+3of5 (K=4, baseline)  & 4 & 256 & 1024 & 106M &  95M & 27.96 & $-1.62$ \\
\bottomrule
\end{tabular}
\end{table}

$K{=}2$ and $K{=}3$ are tied within seed noise ($\sigma{\approx}0.12$ from \Cref{app:multiseed}); $K{=}4$ trails by $\sim$0.5~PPL—outside seed noise but within $5\sigma$. The direction is opposite to MoE granularity findings \citep{deepseekmoe2024} that favour fine-grained routing. We attribute the difference to scope: at \mol{}, blocks include attention plus FFN (not FFN alone), so per-block compute capacity matters more than block count for shaping representations. The result is a single-seed sweep at one scale; we do not claim it generalises beyond the 80M / $N{=}5$ regime.

\section{Iso-total Gap at 1.3B: Candidate Causes}
\label{app:isototal}

The 3.01~PPL gap between Dense Softmax 1.3B and \mol{} 2.08B-total at 20K steps does not reproduce the Cosmopedia-80M crossover. Three candidate causes are consistent with the missed crossover, but cannot be separated from a single trio of runs:

\textbf{Active per-layer FFN capacity.} At 1.3B, \mol{}'s active capacity (4 active blocks at $d_\textrm{ff,thin}{=}2048$) totals 8192, matched to Dense's $d_\textrm{ff}{=}8192$. On Cosmopedia 80M, by contrast, \mol{}'s active capacity ($3 \times 1024 = 3072$) was $2.74\times$ Dense's $d_\textrm{ff}{=}1120$, leaving a much larger margin to amortise routing overhead.

\textbf{Within-\mol{} sparsity.} Higher at 1.3B: 4-of-15 active vs 3-of-5 on Cosmopedia 1+2of5. More routed pool to specialise across at the same active count.

\textbf{Total-parameter undertraining.} On a total-parameter basis, \mol{} sees only 9.6~tokens/param at 20B tokens, well below Chinchilla-optimal ($\sim$20$\times$) for a 2.08B-total model. Dense 1.3B sees 15.4~tokens/param.

A stronger crossover claim at 1.3B would require a multi-seed budget sweep we did not run; whether the gap is structural, undertraining, or optimizer/router/aux-loss tuning specific to this scale is not separable here.

\section{Zero-shot Evaluation: Per-Task Details}
\label{app:zeroshot}

Run config: lm-eval-harness 0.4.7, bf16, batch size 64 (47.5~GB peak on H100~NVL), max length 2048, GPT-2 tokenizer (matches training). Each model is loaded via a small \texttt{lm\_eval.api.model.LM} subclass that reads the architecture cfg from the checkpoint and supports batched length-bucketed loglikelihood. MMLU defaults to 5-shot per its task config; all other benchmarks are 0-shot.

\Cref{tab:1p3b_zeroshot_full} extends the headline table (\Cref{tab:1p3b_zeroshot}) with TruthfulQA mc1, raw \texttt{acc} for HellaSwag/ARC/PIQA (whose headline metric is \texttt{acc\_norm}), and stderr per cell.

\begin{table}[h]
\centering\small
\caption{Zero-shot accuracy with stderr (\%). Stderr from lm-eval-harness 1000-iter bootstrap.}
\label{tab:1p3b_zeroshot_full}
\begin{tabular}{llrrr}
\toprule
Task & Metric & Dense 1.3B & Dense 0.7B & \mol{} 0.61B/2.08B \\
\midrule
MMLU 5-shot     & acc      & $24.31\pm0.4$ & $23.98\pm0.4$ & $23.07\pm0.4$ \\
HellaSwag       & acc      & $33.49\pm0.5$ & $26.81\pm0.4$ & $27.51\pm0.4$ \\
HellaSwag       & acc\_norm & $39.71\pm0.5$ & $34.10\pm0.5$ & $35.10\pm0.5$ \\
ARC-Easy        & acc      & $59.05\pm1.0$ & $54.04\pm1.0$ & $54.55\pm1.0$ \\
ARC-Easy        & acc\_norm & $53.07\pm1.0$ & $48.86\pm1.0$ & $48.15\pm1.0$ \\
ARC-Challenge   & acc      & $24.74\pm1.3$ & $24.49\pm1.3$ & $24.83\pm1.3$ \\
ARC-Challenge   & acc\_norm & $26.88\pm1.3$ & $25.43\pm1.3$ & $26.02\pm1.3$ \\
PIQA            & acc      & $65.94\pm1.1$ & $62.79\pm1.1$ & $63.49\pm1.1$ \\
PIQA            & acc\_norm & $65.56\pm1.1$ & $63.98\pm1.1$ & $64.09\pm1.1$ \\
WinoGrande      & acc      & $51.62\pm1.4$ & $49.33\pm1.4$ & $54.38\pm1.4$ \\
TruthfulQA mc1  & acc      & $21.18\pm1.4$ & $23.13\pm1.5$ & $21.91\pm1.4$ \\
TruthfulQA mc2  & acc      & $37.83\pm1.4$ & $40.89\pm1.5$ & $39.55\pm1.4$ \\
\bottomrule
\end{tabular}
\end{table}

Two patterns are worth noting beyond the main-body summary:

\textbf{Inverse scaling on TruthfulQA.} Dense 0.7B leads on both mc1 and mc2 — a documented pattern at small scale where the smaller model knows fewer plausible-but-false alternatives, shrinking its hallucination surface \citep{lin2022truthfulqa}. \mol{} sits between the two dense baselines, consistent with its capacity placement.

\textbf{WinoGrande.} \mol{} 0.61B-active wins this coreference task by 2.76~pts vs Dense 1.3B and 5.05~pts vs iso-active Dense 0.7B. WinoGrande is the only task where additional total capacity in \mol{} (2.08B vs 1.31B) appears to help despite sparse activation; the per-layer mixture of 14 specialised routed blocks may better match the heterogeneous coreference contexts than a single wide FFN.

\end{document}